# Localization and Navigation System for Indoor Mobile Robot

Yanbaihui Liu[1, *]

[1]Department of Electrical & Computer Engineering, University of Michigan, Ann Arbor, U.S.

*Corresponding author: yanbhliu@umich.edu

**Abstract.** Visually impaired people usually find it hard to travel independently in many public places such as airports and shopping malls due to the problems of obstacle avoidance and guidance to the desired location. Therefore, in the highly dynamic indoor environment, how to improve indoor navigation robot localization and navigation accuracy so that they guide the visually impaired well becomes a problem. One way is to use visual SLAM. However, typical visual SLAM either assumes a static environment, which may lead to less accurate results in dynamic environments or assumes that the targets are all dynamic and removes all the feature points above, sacrificing computational speed to a large extent with the available computational power. This paper seeks to explore marginal localization and navigation systems for indoor navigation robotics. The proposed system is designed to improve localization and navigation accuracy in highly dynamic environments by identifying and tracking potentially moving objects and using vector field histograms for local path planning and obstacle avoidance. The system has been tested on a public indoor RGB-D dataset, and the results show that the new system improves accuracy and robustness while reducing computation time in highly dynamic indoor scenes.

**Keywords:** Visual SLAM; localization; navigation; indoor robot.

## 1. Introduction

Indoor navigation can be challenging for visually impaired individuals, who often struggle with obstacle avoidance and finding their way to their desired destination in indoor public places like airports and shopping malls. Traditional methods of obstacle avoidance and guidance, such as using a cane or a human guide, can be cumbersome and limiting. With the development of time and technology, indoor navigation robots have emerged as a potential solution. However, how to ensure walking safety and localization accuracy in indoor environments where GPS signals are not reliable has become a problem. A popular approach is Visual Simultaneous Localization and Mapping (SLAM) which leverages 3D vision to estimate robot motion and perform mapping when neither the environment nor the location of the sensor is known. Applications of stereo or RGB-D cameras have further improved the robustness and accuracy of Visual SLAM systems and, as a result, are a good way to perform indoor localization and real-time path planning.

There are many existing methods for the navigation system. Blochliger et al. convert sparse feature-based maps from vision SLAM systems to 3D topological maps and use optimized Rapidly-exploring random trees (RRT*) for path planning [1]. Ren et al. applied A* path planning algorithm to laser SLAM to achieve indoor navigation [2]. Long and He proposed deep reinforcement learning-based path planning algorithms that can optimize the decision-making of robots in complex environments [3]. Unfortunately, none of these algorithms can navigate in a highly dynamic environment via visual SLAM, but simply combining classical visual SLAM with SOTA path planning algorithms is also insufficient in navigating modern indoor scenarios due to the limitations of visual SLAM itself. ORB-SLAM2 detects loops and re-localizes the camera given the assumption of a static environment [4]. However, initializing and beginning tracking the sensor pose relative to the first frame causes inaccuracy when the algorithms detect the feature points of dynamic objects. DynaSLAM [5] is built on ORB-SLAM2 but uses Mask-RCNN [6] to detect and segment potential dynamic content from the scene, remove all key points on potential dynamic targets, and paint the vacant background with information obtained from the previous static scene, which able to react to changes produced by static objects at that precise moment the image is captured and works in dynamic environments. The background occluded by dynamic objects is inpainted given the camera positions (output by the tracking and mapping step). However, Mask R-CNN is a complex region-based convolutional neural network (CNN) that segments the image into regions of interest and then sends each region to the CNN, so is computationally costly. Without a powerful GPU, it may be difficult for a robot to use a combination of Mask-RCNN and visual SLAM algorithms in real-time for indoor navigation tasks. Besides, since public spaces like modern offices are mixed with a large number of potential targets that are not moving, such as sitting people, there might be too few feature points left on the map to construct a reliable reference system for future detection and inpainting tasks.

This paper proposes a front-end system combining dynamic environment visual SLAM (DEV-SLAM) and path planning to improve the robustness of indoor guided robot localization and navigation systems while enabling real-time detection to serve as a frontrunner to provide guidance to the user. In such a scenario, the robot must make decisions about handling a large number of potential moving targets when building a map and avoid hitting people or obstacles during path planning. The system removes the actual moving objects and considers static objects as background to improve accuracy, uses a relatively simple structured algorithm for object detection and segmentation to reduce processing time when localizing and mapping, and applies vector field histograms (VFH+) on the rough estimated path for local navigation and obstacle avoidance. The experiment performed on the TUM benchmark dataset shows that the proposed approach successfully eliminates feature points extracted from potentially dynamic objects and achieves faster computational speed. The test run results also show that the new system can properly navigate to the endpoint without hitting obstacles.

## 2. Method

### 2.1 Localization and Mapping

#### 2.1.1 System Overview

The overall procedure of the proposed DEV-SLAM is shown in Fig 1. The algorithm builds on DynaSLAM but accomplishes the detection of potentially dynamic objects with a less computationally intensive alternative method. For each image, segmentation is performed using a pre-trained YOLOv4 model, any feature point within a bounding box labeled as person is the target, then an initial camera trajectory is computed. For each of the bounding boxes, a sparse optical flow analysis is performed with respect to the previous frame and compared to the initial camera trajectory. If there is a sufficient number of optical flow points that deviate from the initial camera trajectory in a given bounding box, that object within the bounding box is considered dynamic and masked out in the image output to the SLAM algorithm, and vice versa.

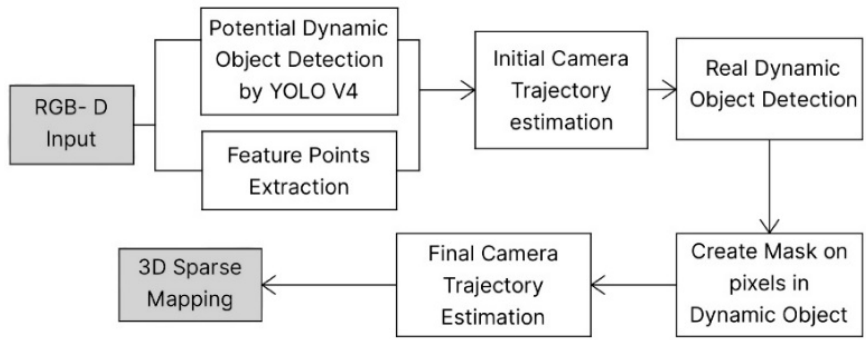

**Fig. 1** Block diagram of DEV-SLAM.

#### 2.1.2 You Only Look Once (YOLOv4)

YOLOv4 is a one-stage object detection model well-known for its exceptional computational speed and accuracy [7]. Unlike Mask R-CNN, YOLO uses CNN once on the entire image to split it into a grid. For each grid location, the CNN network generates a fixed number of different bounding boxes and computes a class probability for each of them. CNN then uses a threshold for the intersections over unions to eliminate bounding boxes that are likely to refer to the same object. This structure of YOLO results in faster computation speed.

**2.1.3 Optical Flow Based Moving Object Detection**

Moving object detection aims to extract changing regions from image sequences based on background images. The mainstream moving object detection methods include frame difference, background difference, and optical flow methods. This paper takes optical Flow based method since it estimates the shift vector for pixels between two consecutive frames generated by the movement of cameras or objects and can reach higher accuracy in complex scenarios and avoid the occlusion and overlapping of objects to the greatest extent possible [8,9].

In general, optical flow can be divided into sparse optical flow and dense optical flow. Sparse optical flow uses a specific set of pixels for estimation, while dense optical flow calculates the displacement vector for each pixel in the image. This paper uses a sparse optical flow implemented by a Sparse Pyramid Lucas-Kanade algorithm to improve the detection speed while avoiding the inaccuracy of object detection caused by lost feature points when the detection interval is comparatively long [10].

**2.1.4 Real Dynamic Object Detection**

The algorithm for dynamic object tracking and segmentation (DOT) is shown in Algo 1. The darknet YOLOv4 model [11] that is pre-trained on the MS COCO [12] is used to segment all potentially moving instances presented in a frame. The instances not affiliated with the target class are directly categorized as the background. For the segmented potentially moving object, the sparse optical flow is used to further determine whether the object is moving.

The OpenCV function goodFeaturesToTrack finds the Shi-Tomasi corner points to track on the previous frame, and then uses the OpenCV function calcOpticalFlowPyrLK that implements the Pyramid Lucas-Kanade method to calculate the optical flow at these points [9]. The output of this algorithm is the relative motion of each point in the camera frame and can be used to decide whether those pixels are moving or stationary. This algorithm uses a tunable threshold metric to determine whether an object is moving or stationary. A threshold on the resultant of the motion of each pixel is used to determine if it's a dynamic pixel and another threshold for the number of dynamic pixels in a bounding box is used to determine if the whole object is moving. Last, for those moving objects, masks are created to remove the feature points within bounding boxes.

The whole bounding box is masked off without delineating the specific contours of the person for two reasons: one is that there is no need for extra consideration of specific dynamic contents, such as

```
Algorithm 1 Dynamic Object Detection
Require: CurFrame, PrevFrame
    boxes ← YOLOv4              ▷ Potential Object Detection
    d ← displacement of feature points in two consequence
    frames
    if d > threshold1 then
        DynamicPts ← Position of feature points in CurFrame
    end if
    while i = 0; i < numBoxes; i++ do
        count ← number of DynamicPts within current bound-
    ing box
        if count > threshold2 then
            Potential object is actually moving.
        end if
    end while
```

the mug held by the person, and two, the impact caused by delineating object boundaries or not is minimal due to the sparsity of feature points. Fig 2 shows an example of different masks created by DynaSLAM and DEV-SLAM in a highly dynamic scene. Fig 3 illustrates the difference between DynaSLAM and DEV-SLAM for stationary persons. DynaSLAM is masking the persons even though they are stationary at the desk, while the DEV-SLAM tracking feature points to the stationary persons.

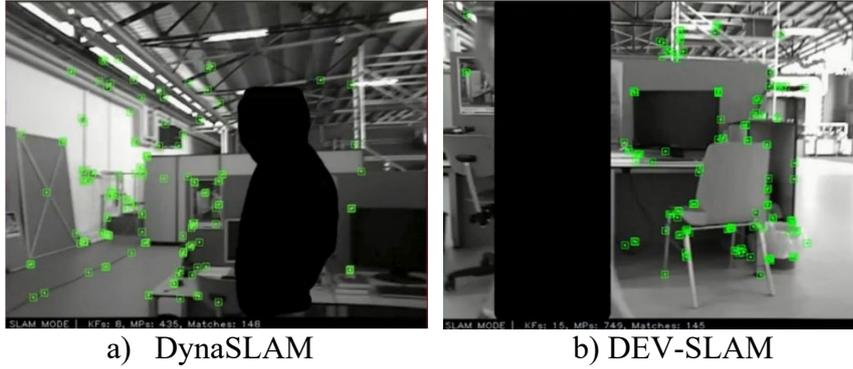
a) DynaSLAM　　　　　　　　b) DEV-SLAM
**Fig. 2** Results of masks and feature points in the walking scenario.

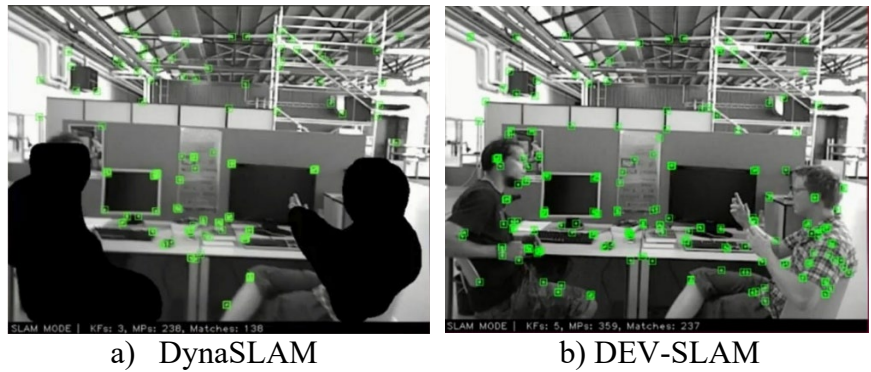
a) DynaSLAM　　　　　　　　b) DEV-SLAM
**Fig. 3** Results of generated masks and feature points in sitting scenario.

**2.2 Navigation**

A* is a classic heuristic search algorithm for solving shortest paths in static road networks [13]. The cost function of the initial state to the target state via state n of A* is

$$f(n) = g(n) + h(n) \qquad (1)$$

where g(n) represents the actual cost from the initial state to state n, while h(n) represents the estimated cost of the optimal path that is usually found by Manhattan distance or Euclidean distance. However, using a* cannot successfully get the path planning job done within a modern office scenario. Moving staff or visitors can turn a static office map into a highly dynamic environment, thus a local path planning algorithm VFH+ is added to achieve obstacle avoidance while optimizing trajectories, otherwise the robot needs to wait for new objects disappear or rescans the whole environment. The essence of VFH+ is to set up a local map coordinate system with the robot as the geometric center and lets the robot choose the direction with the smallest value and the largest width to advance based on the created polar histogram [14]. For creating the histogram, it first converts the obstacle vector for each cell in the circular active window into a polar obstacle density, which is then represented as a binary map using a hysteresis bithreshold, and finally, it predicts the trajectory based on the current motion state (v, w), searches for the angle of the left and right boundary cells that have overlap with the trajectory, determines the left and right boundaries of feasible sectors, and updates the Masked polar histogram. When the robot is far away from the target point on the planned path, the density is small, and the robot can continue to approach the target point. When the density is increased to a pretty large number, global planning intervenes.

## 3. Experiments

**3.1 Localization and Mapping**

The proposed SLAM system is evaluated with the SOTA 3D object detection methods ORB-SLAM2 and DynaSLAM in the camera frame and compared the results of Absolute Trajectory Error (ATE), which measures the average deviation from ground truth trajectory per frame, and Relative Position Error (RPE), which measures the quality of localization [15]. The 3D trajectories of models are also given to compare each model visually. Besides the SOTA method, a performance of DynaSLAM (YOLO), a method that naively replaces the time-consuming Mask R-CNN model with YOLOv4, is used as a bridge to connect several models. Overall performance is tested and obtained on an Intel® Core™ i9-11900H Processor, and the number of feature points is set to 1250.

### 3.1.1 Dataset

This paper adopts a pre-trained COCO Instance Segmentation baseline model with Mask R-CNN R50-FPN 3x [16] and Darknet pre-trained model with YOLOv4. Experiments are performed on 'freiburg3_walking_xyz' and 'freiburg3_sitting_xyz' subsets of the TUM RGB-D [17] dataset that is composed of RGB images, depth images, accelerometer data, and ground truth for the trajectory of the camera. 'freiburg3_walking_xyz' includes two people walking around a desk that the sensor is being translated along with the 3 directions is used to evaluate the robustness of the algorithm to moving objects in indoor scenes. 'freiburg3_sitting_xyz' that includes two people sitting at a desk with slight movements of their torso and hands while the sensor is being translated along with the 3 directions is used to assess whether the algorithm could correctly identify the actual moving objects.

### 3.1.2 Trajectories

Fig 4 and Fig 5 show the estimated trajectories of DEV-SLAM, DynaSLAM, and ORB-SLAM2 and compared with the ground-truth on the walking and sitting test sets. While both DEV-SLAM and DynaSLAM do not perfectly fit the ground truth, we can still virtually see they are improved a lot from the ORB-SLAM2 on the walking scenario. ORB-SLAM2 is more sensitive to environmental changes and leads to significant deviations or sometimes lack of ability to track a certain dimension. The difference in the performance of the sitting scenario is difficult to show directly through trajectory due to the significant improvement in ORB-SLAM2 caused by more static scenes. Models with masked moving objects can track camera poses better in more dynamic scenes.

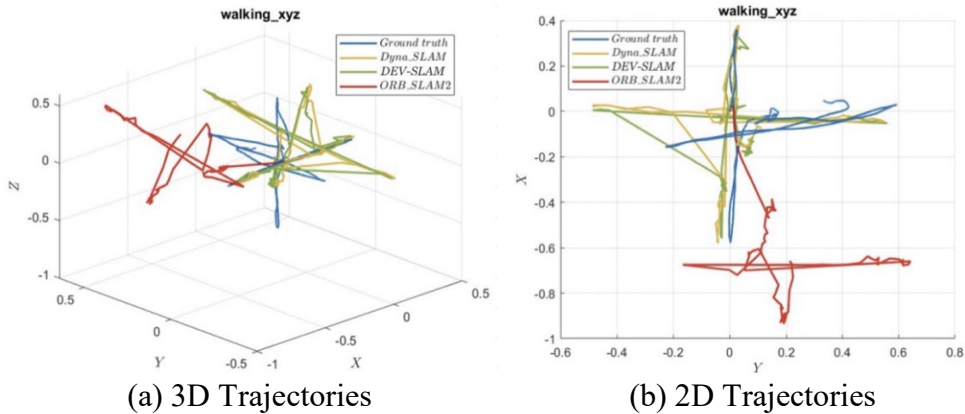

(a) 3D Trajectories      (b) 2D Trajectories
**Fig. 4** Trajectories of different models on freiburg3_walking_xyz dataset.

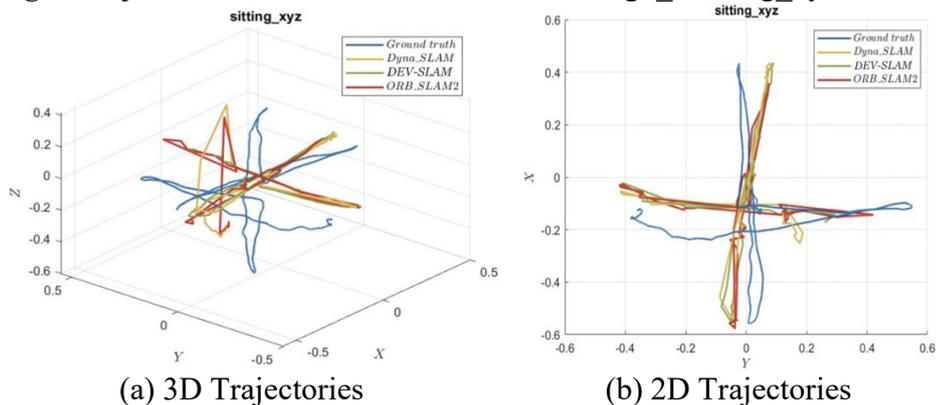

(a) 3D Trajectories      (b) 2D Trajectories
**Fig. 5** Trajectories of different models on freiburg3_sitting_xyz dataset.

### 3.1.3 Evaluations Metrics

The Relative Pose Error (RPE) mainly describes the accuracy of the positional difference between two frames separated by a fixed time difference Δ (compared to the true positional), which is equivalent to a direct odometer measurement error. Defining algorithm estimated poses as $P_1,...,P_n \in SE(3)$ ground-truth poses as $Q_1,...,Q_n \in SE(3)$, where n is the number of frames, and Δ as the time interval, the RPE of the i-th frame $E_i$ can be defined as:

$$E_i := (Q_i^{-1}Q_{i+\Delta})(P_i^{-1}P_{i+\Delta}) \qquad (2)$$

Knowing the total number n with interval Δ, we can get m: = n - ΔRPE, then the root mean square error (RMSE) can be calculated by:

$$RMSE(E_{1:n}, \Delta) := \left(\frac{1}{m}\sum_{i=1}^{m}||trans(E_i)||^2\right)^{\frac{1}{2}} \qquad (3)$$

where trans(·) stands for taking the translation. Table 1 shows RPE comparison results on walking and sitting sequences against RGB-D ORB-SLAM2, and DynaSLAM. In a highly dynamic scenario, even though DEV-SLAM has larger translational and rotational errors than DynaSLAM, the numerical difference appears in the thousandth percentile, which is relatively small compared to how much we outperform ORB-SLAM2. For static scenarios, in terms of camera translation, DEV-SLAM outperforms DynaSLAM and is closer to the results of ORB-SLAM2; in terms of camera rotation, DEV-SLAM has the lowest error among methods.

The absolute trajectory error (ATE) measures the difference between the estimated pose and ground-truth poses, which can intuitively reflect the algorithmic accuracy and global consistency of the trajectory. ATE of the i-th frame $F_i$ can be defined as:

$$F_i := Q_i^{-1}SP_i \qquad (4)$$

where $S \in SE(3)$ is a transformation matrix from the estimated pose to the real pose calculated by the least squares method, and similar to RPE, RMSE of ATE can be defined as:

$$RMSE(F_{1:n}, \Delta) := \left(\frac{1}{m}\sum_{i=1}^{m}||trans(F_i)||^2\right)^{\frac{1}{2}} \qquad (5)$$

**Table 1**. Mean and RMSE of RPE for RGB-D Cameras

| Type | translational error | | | rotational error | | | dataset |
|---|---|---|---|---|---|---|---|
| | RMSE (m) | mean (m) | median (m) | RMSE (deg) | mean (deg) | median (deg) | |
| DynaSLAM | 0.02780 | 0.02477 | 0.02323 | 0.59822 | 0.53835 | 0.50755 | walking_xyz |
| DynaSLAM (YOLO) | 0.03051 | 0.02664 | 0.02546 | 0.61130 | 0.54036 | 0.49596 | |
| DEV-SLAM | 0.02787 | 0.02566 | 0.02215 | 0.60830 | 0.58746 | 0.55259 | |
| ORB-SLAM2 | 0.54071 | 0.44722 | 0.38745 | 10.45796 | 8.56084 | 7.29018 | |
| DynaSLAM | 0.02326 | 0.02085 | 0.01988 | 0.74210 | 0.64352 | 0.57356 | sitting_xyz |
| DynaSLAM (YOLO) | 0.02382 | 0.02112 | 0.01965 | 0.57789 | 0.51704 | 0.48900 | |
| DEV-SLAM | 0.02238 | 0.01873 | 0.01611 | 0.45975 | 0.40283 | 0.36641 | |
| ORB-SLAM2 | 0.02049 | 0.01809 | 0.01653 | 0.59524 | 0.53073 | 0.49641 | |

Table 2 shows ATE results on the same sequences. However, DEV-SLAM outperforms ORB SLAM2 and reaches the same level as DynaSLAM in a highly dynamic situation. In a relatively static case, our method beats DynaSLAM by 7% but is 6% lower than the ORB-SLAM2. The performance on the sitting sequence is worse than ORB-SLAM2 because DEV-SLAM inherits some of the problems of masking out the entire bounding box of dynamic objects thus removing some useful information on backgrounds. Also, due to some uncontrollable errors, the proposed method still has the possibility of mistaking static objects for dynamic ones.

Table 2. Mean and RMSE of ATE for RGB-D Cameras

| Type | RMSE (m) | mean (m) | median (m) | dataset |
|---|---|---|---|---|
| DynaSLAM | 0.01950 | 0.01707 | 0.01583 | Walking_xyz |
| DynaSLAM (YOLO) | 0.02210 | 0.01893 | 0.01661 | |
| DEV-SLAM | 0.01955 | 0.01759 | 0.01393 | |
| ORB-SLAM2 | 0.19732 | 0.17668 | 0.15489 | |
| DynaSLAM | 0.01601 | 0.01460 | 0.01441 | sitting_xyz |
| DynaSLAM (YOLO) | 0.01523 | 0.01369 | 0.01266 | |
| DEV-SLAM | 0.01501 | 0.01245 | 0.00920 | |
| ORB-SLAM2 | 0.01415 | 0.01210 | 0.01035 | |

The frame rates measured in frames per second (fps) are shown in Table 3. By comparing DynaSLAM (YOLO) with DEV-SLAM, even though an extra DOT module is added, the overall detection speed does not change much and still outperforms DynaSLAM; in scenes with many static potential targets, the overall frame rates even improved 7%.

**3.2 Navigation**

The experiment verifying the successful navigation and obstacle avoidance rate is performed on an ROS-simulated environment as shown in Fig 6. The robot is supported to start from its initial location to reach the target point (shown as a blue point in the plot). In this scene, there are boxes, wooden blocks and other things as obstacles, the initial position of the robot is the origin, and the expected arrival position is the final coordinate. The experiment requires the robot to be able to reach the target point without collision. The experiment proves that the robot can basically complete the task, but if the obstacle is set very close to the target point, the probability of failure increases a lot.

Table 3. FPS for RGB-D Cameras

| Type | Median FPS | Mean FPS | dataset |
|---|---|---|---|
| DynaSLAM | 0.2306 | 0.2154 | walking_xyz |
| DynaSLAM (YOLO) | 2.8221 | 1.6958 | |
| DEV-SLAM | 3.0823 | 2.0254 | |
| ORB-SLAM2 | 59.3521 | 53.9689 | |
| DynaSLAM | 0.2301 | 0.2099 | sitting_xyz |
| DynaSLAM (YOLO) | 2.8399 | 2.7998 | |
| DEV-SLAM | 3.0630 | 2.7748 | |
| ORB-SLAM2 | 52.9967 | 49.1688 | |

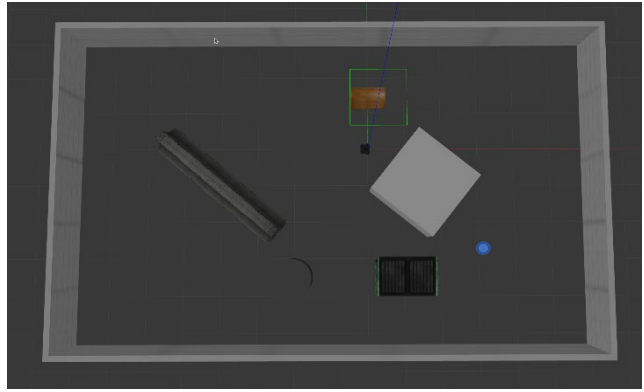

**Fig. 6** ROS world.

## 4. Summary

This paper presented a front-end indoor navigation system for visually impaired users that robustly does localization and mapping in both static and dynamic environments by combining dynamic object detection and visual SLAM and avoiding obstacles to reach the desired location. Using YOLOv4 to segment potentially dynamic objects is a promising alternative to the computationally intensive Mask R-CNN network for masking out dynamic objects before they are used in a visual SLAM algorithm. For a small compromise on performance compared with using Mask-RCNN, this approach promises significantly faster execution speed. Based on the localization and local path planning the visually impaired users can be successfully guided to the desired location without hitting. However, the path planning method this paper is using now does not guarantee the successful rate when the obstacle is close to the target location, which can be one of the future works. Also, how to furtherly increase the detection and mapping speed so that it can approach the same level as ORB-SLAM2 is also a point to consider.